\PassOptionsToPackage{prologue,dvipsnames}{xcolor}
\documentclass[sigconf, screen]{acmart}

\usepackage{graphicx}
\usepackage{amsmath}
\usepackage{booktabs}
\usepackage{amsfonts}
\usepackage{mathrsfs}
\usepackage{amsthm}
\usepackage{multirow}
\usepackage{cuted}
\usepackage{subfiles}
\usepackage{color}

\newcommand{\todo}{\textcolor{black}}
\newcommand{\wj}{\textcolor{black}}

\newcommand{\new}{\textcolor{black}}
\newcommand{\best}{\textcolor{red}}
\newcommand{\second}{\textcolor{blue}}

\newcommand{\etal}{\textit{et al}.}
\newcommand{\ie}{\textit{i}.\textit{e}.}
\newcommand{\eg}{\textit{e}.\textit{g}.}

\AtBeginDocument{%
  }

\copyrightyear{2022}
\acmYear{2022}
\setcopyright{acmcopyright}\acmConference[MM '22]{Proceedings of the 30th ACM International Conference on Multimedia}{October 10--14, 2022}{Lisboa, Portugal}
\acmBooktitle{Proceedings of the 30th ACM International Conference on Multimedia (MM '22), October 10--14, 2022, Lisboa, Portugal}
\acmPrice{15.00}
\acmDOI{10.1145/3503161.3547991}
\acmISBN{978-1-4503-9203-7/22/10}

\begin{document}

\title{Self-Aligned Concave Curve: Illumination Enhancement for Unsupervised Adaptation}


\author{Wenjing Wang}
\affiliation{%
  \institution{Peking University}
  \city{Beijing}
  \country{China}
}
\email{daooshee@pku.edu.cn}

\author{Zhengbo Xu}
\affiliation{%
  \institution{Peking University}
  \city{Beijing}
  \country{China}
}
\email{icey.x@pku.edu.cn}

\author{Haofeng Huang}
\affiliation{%
  \institution{Peking University}
  \city{Beijing}
  \country{China}
}
\email{hhf@pku.edu.cn}

\author{Jiaying Liu}

\authornote{Corresponding Author. This work was supported by the National Natural Science Foundation of China under Contract No.62172020, and a research achievement of Key Laboratory of Science, Techonology and Standard in Press Industry (Key Laboratory of Intelligent Press Media Technology).}
\affiliation{%
  \institution{Peking University}
  \city{Beijing}
  \country{China}
}
\email{liujiaying@pku.edu.cn}

\renewcommand{\shortauthors}{Wenjing Wang, Zhengbo Xu, Haofeng Huang, \& Jiaying Liu}


\begin{abstract}
Low light conditions not only degrade human visual experience, but also reduce \wj{the performance of downstream machine analytics}.
Although \wj{many} works have been designed for low-light enhancement or domain adaptive machine analytics, the former considers less on high-level vision, while the latter neglects the potential of image-level signal adjustment.
\wj{How to restore underexposed images/videos from the perspective of machine vision has long been overlooked.}
In this paper, we are the first to propose a learnable illumination enhancement model for high-level vision.
Inspired by real camera response functions, we assume that the illumination enhancement function should be a concave curve, and \wj{propose} to satisfy this concavity \wj{through discrete integral.}
With the intention of \wj{adapting illumination} from the perspective of machine vision \wj{without task-specific annotated data}, we design an asymmetric cross-domain self-supervised training strategy.
Our model architecture and training designs mutually benefit each other, forming a powerful unsupervised normal-to-low light adaptation framework.
\wj{Comprehensive} experiments demonstrate that our method surpasses existing low-light enhancement and adaptation methods and shows superior generalization on various low-light vision tasks, \wj{including classification, detection, action recognition, and optical flow estimation.}
All \wj{of} our data, code, and results will be available online upon publication of the paper.
\end{abstract}



\maketitle

\vspace{-2mm}
\section{Introduction}
\label{sec:intro}

Insufficient lighting is a common kind of image degradation, caused by dark environments, corrupted equipment, or improper shooting settings.
It may degrade the visual quality of images, causing low visibility, lost details, and aesthetic distortion.
Besides, with the rise of machine learning, visual analytics has been playing an increasingly critical role in many applications.
Low light can also pose threats to machine analytics, presenting challenges to high-level vision tasks in the low-light condition, such as nighttime autonomous driving and surveillance video analysis.



The research of restoring low-light images has attracted wide attention.
From early manually designed algorithms~\cite{Enhance_MSRCR} to recent data-driven deep models~\cite{RUAS}, a great number of works \wj{have effectively improved} human visual quality of low-light images.
However, most existing low-light enhancement methods \wj{do not take machine vision into account}.
Some methods introduce noise removal~\cite{Enhance_RetinexNet} or detail reconstruction~\cite{Enhance_CVPR20}, which enhances visual quality but might mislead high-level analytics models.
Some other methods take semantic perception into consideration~\cite{HDR_LiuLCK0CH20}, but they still focus on human vision and have unsatisfactory performance on downstream high-level vision tasks.

In recent years, to promote the development of machine analytics in low-light scenarios, many datasets have been built~\cite{DARKFACE,FA_ExDark_CVIU19}, giving birth to a series of machine vision models tailored for underlit environments.
With a large amount of training data, many methods improve normal light frameworks by incorporating the properties of insufficient lighting conditions~\cite{MAET}.
Some works further explore the task of adapting normal light models to low-light without task-specific annotation, which is more universal and application-driven~\cite{DANNet,HLAFace,CIConv}.
However, existing normal-to-low light adaptation algorithms \wj{either} rely on multiple sources~\cite{DANNet}, adopt troublesome multi-stage and multi-level processes~\cite{HLAFace}, or fail in darker cases~\cite{CIConv}.
Moreover, most adaptive methods concentrate on the high-dimensional features of machine analytics models, but neglect the characteristics of input images themselves.

Different from existing normal-to-low light adaptation methods, we make full use of the potential of illumination adjustment.
\wj{We first build an illumination enhancement model which can maximize the model's abilities while being easy to learn.}
Inspired by camera response functions, we design a new \wj{model paradigm}: \textit{deep concave curve}, which can determine the new pixel value in the enhanced result with a high degree of freedom.
To effectively satisfy concavity, we propose to first predict a non-positive second derivative, then apply discrete integral implemented by convolutions.
\wj{To train} this model towards unsupervised adaptation, we design \textit{asymmetric self-supervised alignment}.
On the normal light side, we learn decision heads with a self-supervised pretext task. Then on the low-light side, we fix the decision heads and let our model improve the pretext task performance through enhancing the input image.
In this way, \wj{even without annotated data}, our model can learn how to make the machine analytics model better perceive the enhanced low-light image.
To make full use of image information and provide good guidance for illumination enhancement, we propose a new rotated jigsaw permutation task.
Experiments show that our model architecture and training design are compatible with each other.
On one hand, our self-learned strategy can better restore illumination compared with other feature adaptation strategies; on the other hand, our deep concave curve can best maximize the potential of self-learned illumination alignment.

The proposed illumination enhancement model, \todo{self-aligned concave curve (SACC)}, can serve as a powerful tool for unsupervised low-light adaptation. Although \todo{SACC} does not require normal or low-light annotations and \wj{does not even} adjust the downstream model, it achieves superior performance on a variety of low-light vision tasks.
To further deal with noises and semantic domain gaps, we propose to adapt downstream analytics models by pseudo labeling.
\wj{Finally, we build an adaptation framework \todo{SACC+}, which is concise and easy to implement but can outperform existing low-light enhancement and adaptation methods by a large margin.}

In summary, our contributions are threefold:
\vspace{-1.5mm}
\begin{itemize}
    \item We are the first to propose a learnable pure illumination enhancement model for high-level vision. Inspired by camera response functions, we design a deep concave curve. Through discrete integral, the concavity constraint can be satisfied through the model architecture itself.
    \item Towards unsupervised normal-to-low light adaptation, we design an asymmetric cross-domain self-supervised training strategy. Guided by the rotated jigsaw permutation pretext task, our curve can adjust illumination from the perspective of machine vision.
    \item 
    To verify the effectiveness of our method, we explore various high-level vision tasks, including classification, detection, action recognition, and optical flow estimation.
    Experiments demonstrate our superiority over both low-light enhancement and adaptation state-of-the-art methods.
\end{itemize}

\vspace{-3mm}
\section{Related Works}
\label{sec:related}

{\flushleft {\bf Low-light Enhancement.}} 
Early methods manually design illumination models and enhancement strategies.
In the Retinex theory~\cite{Land1977TheRT}, images are decomposed into reflectance (albedo) and shading (illumination).
On this basis, Retinex-based methods~\cite{Enhance_LIME,Enhance_MF} first decompose images and then \wj{either} separately or simultaneously process the two components.
Histogram equalization and its variants~\cite{Enhance_HE} instead redistribute the intensities on the histogram.

Recent methods are mainly based on deep learning.
Some mod-

\noindent els mimic the Retinex decomposition process~\cite{Enhance_RetinexNet,Enhance_KinD}.
RUAS~\cite{RUAS} unrolls the optimization process of Retinex-inspired models and searches desired network architectures.
EnlightenGAN~\cite{Enhance_EnlightenGAN} introduces adversarial learning.
Zero-DCE~\cite{Enhance_ZeroDCE} designs a curve-based low-light enhancement model and learns in a zero-reference way.
Some methods also target RAW images~\cite{Enhance_SID}, videos~\cite{SID_Motion,SMOID}, and introduce extra light sources~\cite{Red_Flash,Flash}.  Interested readers may refer to \cite{IJCV_survey} and \cite{TPAMI_survey} for comprehensive surveys.

Existing low-light enhancement methods \wj{disregard} downstream machine learning tasks. In comparison, our model targets high-level vision and \wj{greatly benefits} machine vision in low-light scenarios.


{\flushleft {\bf High-Level Vision in Low-light scenarios.}} 
With \wj{an} increasing demand for autonomous driving and surveillance analysis, low-light high-level vision has attracted \wj{ever-higher} attention in recent years.
For dark object detection, Sasagawa~\etal~\cite{FA_YOLODARK_ECCV20} merged pretrained models in different domains with glue layers and generative knowledge distillation. 
MAET~\cite{MAET} learns through jointly decoding degrading transformations and detection predictions.
HLA-Face~\cite{HLAFace} adopts a joint pixel-level and feature-level adaptation framework.
For nighttime semantic segmentation, DANNet~\cite{DANNet} employs adversarial training to adapt models in one stage without additional day-night image transferring.
For general tasks, CIConv~\cite{CIConv} designed a color invariant representation.
Some works also focus on tasks of image retrieval~\cite{Retrieval}, depth estimation~\cite{Depth}, and matching~\cite{Matching} in low-light conditions.

\wj{Despite all these progress on} high-level vision in low-light scenarios, many methods rely on low-light annotations, which are \wj{neither} robust \wj{nor} flexible enough.
Existing unsupervised adaptation methods focus on feature migration and ignore the importance of pixel-level adjustment.
Based on illumination enhancement, we propose a new method for low-light adaptation that outperforms existing methods by a wide margin.


\vspace{-0.5mm}
\section{Deep Concave Curve}
\label{sec:curve}

In this section, we introduce the motivation and detailed architecture of our illumination enhancement model.

\vspace{-1.5mm}
\subsection{From CRF to Concave Curve}

Digital photographic cameras use camera response functions (CRFs) \wj{when} mapping irradiance to intensities.
\wj{Although scene illumination changes linearly on the irradiance level, to} fit the logarithmic perception of human vision, cameras employ non-linear CRFs, making illumination adjustment complicated on the intensity level.
To utilize the linearity of irradiance, \wj{some low-light enhancement methods~\cite{HuangYH020,huangtip2022}} transform intensities to irradiance, adjust the irradiance, and then map irradiance back to intensities.
However, back-and-forth irradiance $\leftrightarrow$ intensity mapping is inconvenient and difficult to introduce high-level machine vision guidance.

We propose to simplify the above complex \wj{pipeline} into one single intensity-level adjustment, which is denoted by $g$. We first analyze what form $g$ should take.
Ignoring spatial variations like lens fall-off~\cite{AsadaAB96}, vignetting, and signal-dependent noise, CRF can be assumed to be the same for each pixel in an image. 
Accordingly, we set $g$ to be spatially shared.
Second, to follow the numerical range of pixels and preserve order, $g$ should pass $(0,0)$, $(1,1)$, and \wj{increase monotonically.}
Additionally, although pixels are discrete, we want $g$ to appear roughly continuous, \ie, like a curve.
\wj{Despite} the above constraints, the solution space of $g$ is still too large, \wj{making optimization difficult.}

\begin{figure}[t]
    \centering
  \includegraphics[width=0.99\linewidth]{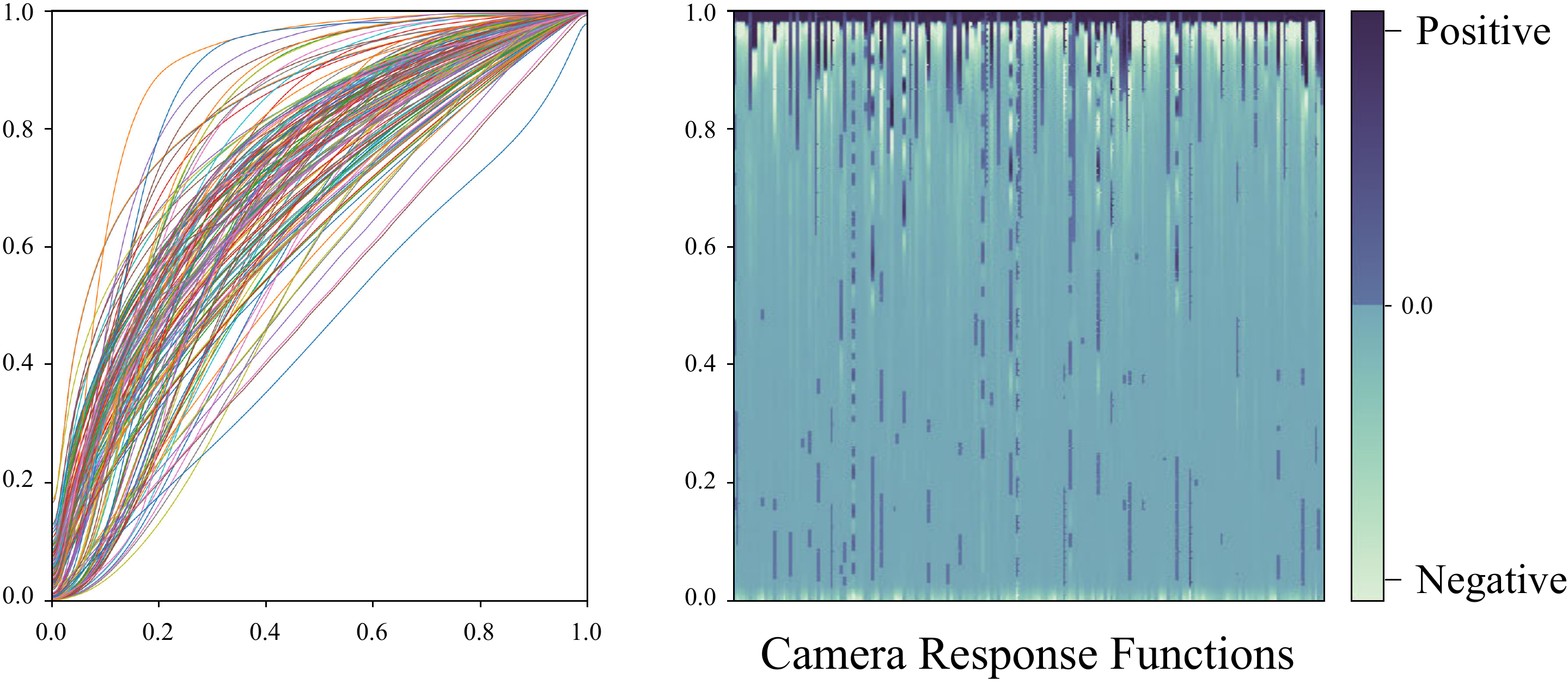}
  \vspace{-1mm}
  \caption{Left: real camera CRFs from the DoRF~\cite{DoRF} dataset. Right: the heat map of second-order derivatives in DoRF. Each column represents a CRF.}
  \label{fig:second_derivative}
  \vspace{-3mm}
\end{figure}

DoRF~\cite{DoRF} is a dataset of real camera CRFs. As shown in Fig.~\ref{fig:second_derivative}, we find that most CRFs are concave. 
Statistically, 89.5\% of the discrete second derivatives are negative. \wj{This is because} human perception of brightness is more sensitive to relative differences between darker tones than differences between lighter tones.
Accordingly, enhancing illumination on irradiance results in concave transformation on intensities.
Therefore, we assume that $g$ should be concave.


With the concavity constraint narrowing the solution space, in Sec.~\ref{sec:learning} we will develop a learning strategy, through which an appropriate $g$ can be automatically found.

\begin{figure*}[t]
    \centering
  \includegraphics[width=0.99\linewidth]{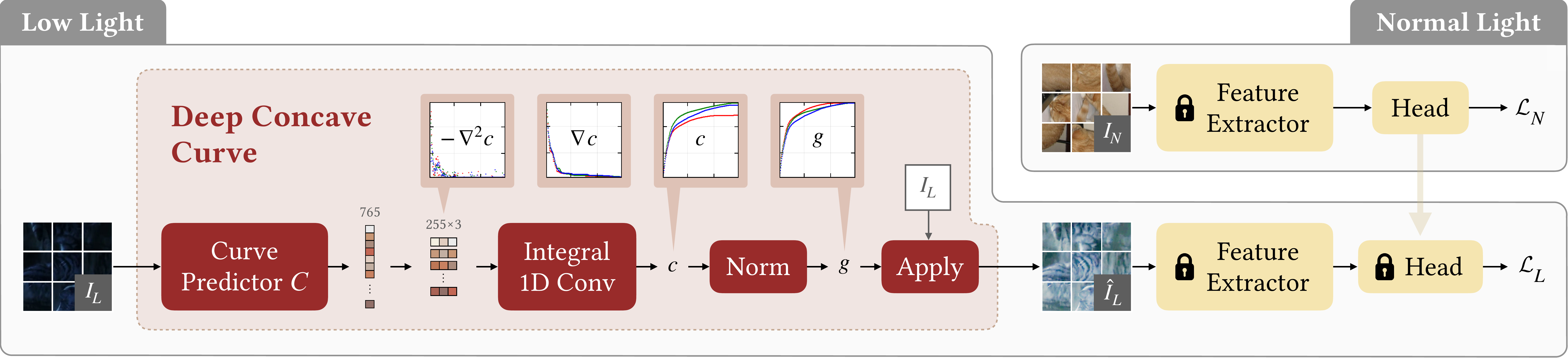}
  \vspace{-1mm}
  \caption{The framework of our model. Our model first predicts a non-negative minus second derivative $-\nabla^2c$, and then integrates and normalizes it into a concave curve $g$, which controls the illumination of the enhanced result. The model is trained in an asymmetric self-supervised way. Based on a pretrained and fixed-weight feature extractor, we first train a pretext head on normal light images, and then train our model with the fixed-weight pretext head on low-light images.}
  \label{fig:framework}
  \vspace{-1.5mm}
\end{figure*}

\vspace{-1.5mm}
\subsection{Form and Implementation}

\todo{Based on the above analysis}, our illumination enhancement adjustment function \wj{should be \textit{monotonically increasing} and \textit{concave}}. The model is called: deep concave curve.

Now we introduce the detailed design of our model. 
We use a neural network $C$ to predict $g$ from the input low-light image $I_L$.
\wj{Our $g$ is a mapping from original pixel values to new pixel values.}
If the color space has $P$ numeric values (\eg, 256 for 8-bit images), then $g$ will be an $\mathbb{R}^P$ vector.
For original pixels of value $(p-1)/P$ on the input image $I_L$, where $p \in \{1,2,..,P\}$, their new pixel values would be the $p$-th element of $g$.
For ease of understanding, we provide an example in the supplementary.

Intuitively, the conditions of monotonic increasing and concavity can be fulfilled by penalty terms on the discrete first and second derivatives.
However, this na\"ive strategy will introduce new losses, adding the burden of balancing multiple learning objectives.

We propose to satisfy these two constraints through the model architecture itself.
Instead of directly estimating $g$, we first predict a minus second derivative $-\nabla ^2 c = C(I_L)$.
By setting the last layer of network $C$ to ReLU, we ensure that $-\nabla ^2 c \geq 0$.
Then, we integrate $-\nabla ^2 c$ into $c$.
Finally, to fit the range of $[0,1]$, we normalize $c$ into $g=\text{Norm}(c)$.
Since $-\nabla ^2 c \geq 0$, $g$ is guaranteed to be concave.

\wj{The discrete integral from $-\nabla ^2 c$ to $\nabla c$ can be implemented as subsequence summation. Denote matrix-vector multiplication $\nabla c = A \times (- \nabla ^2 c)$, we set $A=[a_{ij}]$ as an upper triangular matrix:
\vspace{-0.5mm}
\begin{align}
    a_{ij} = \left\{
        \begin{aligned}
        1,  && i \leq j,\\
        0,  && i > j.
        \end{aligned}
    \right.
\vspace{-0.5mm}
\end{align}
Since $-\nabla ^2 c \geq 0$ and $a_{ij} \geq 0$, we can derive that $\nabla c \geq 0$, thus $g$ is guaranteed to \wj{increase monotonically}.
Similarly, denote $c = B \times (\nabla c)$, we set $B=[b_{ij}]$ as a strictly lower triangular matrix:
\begin{align}
\vspace{-0.5mm}
    b_{ij} = \left\{
        \begin{aligned}
        0,  && i \leq j, \\
        1,  && i > j.
        \end{aligned}
    \right.
\vspace{-0.5mm}
\end{align}
For simplicity,} we merge the two integral into one step $c = D \times (- \nabla ^2 c)$, where $D = B \times A$ is pre-calculated.
Since the first element of $g$ is known to be zero, we only need to predict the \wj{other} $P-1$ values, \ie, $-\nabla ^2 c \in \mathbb{R}^{P-1}$ and $c \in \mathbb{R}^{P}$. 
Finally, for normalization, we divide the vector by its maximum value: $g = \text{Norm}(c) = c / ||c||_{\infty}$.

\vspace{-1.5mm}
\subsection{Network Architecture}

We predict \wj{an} independent $g$ for each color channel.
For RGB images, we have $g_{R}$, $g_{G}$, $g_{B}$.
Since different color channels share the same summation matrix, we put $g_{R}$, $g_{G}$, and $g_{B}$ side-by-side to form a 1D tensor, and implement integral as a 1D convolution with 256 kernels of size 1.
The convolution weight is fixed to $D = B \times A$.

The architecture is illustrated in the red part of Fig.~\ref{fig:framework}.
\wj{To increase} receptive field and acceleration, we first down-sample the input image to $16\times16$ resolution.
Curve predictor $C$ consists of a shallow U-net~\cite{UNet}, two convolutions, and one fully connected layer. 
For 8-bit RGB, the output dimension is 765.
Then, we reshape the output into 255$\times$3, and feed it to the integral 1D convolution.
Finally, \wj{we apply $g$ on the image $I_L$, which is} efficiently implemented with the \textsf{index\_select} function in PyTorch~\cite{PyTorch}.

\section{Asymmetric Self-Supervised Alignment}
\label{sec:learning}

In the previous section, we \wj{have proposed} the deep concave curve for illumination enhancement. In this section, we \wj{will} introduce how to train it towards machine vision.

\vspace{-1.5mm}
\subsection{How to Guide Illumination Enhancement}

Our goal is to align the brightness of low-light images to normal light and improve the performance of downstream tasks in low-light conditions.
The main idea is to employ high-level vision models as guidance.
We first pretrain downstream models on normal light images, then fix their backbones and let our deep concave curve bridge the feature-level gap between low and normal light.

The challenge is that our deep concave curve can only adjust illumination, namely signal \wj{restoration, but} low and normal light also differs in semantics. 
To narrow feature-level gaps, existing methods adopt discrepancy metrics~\cite{DeepCORAL} and adversarial learning~\cite{DANN}.
However, discrepancy metrics and adversarial discriminators \wj{bring} extra semantic supervision, which can mislead our model and make training hard to converge.

To disentangle signal \wj{restoration} and semantics, we adopt cross-domain self-supervised pretext tasks. 
The framework is shown in Fig.~\ref{fig:framework}.
We first train normal light self-supervised learning with \wj{a} fixed-weight feature backbone and a trainable multilayer perceptrons (MLP) head. 
Although enhanced low-light data and normal light data may contain different objects and scenes, they all follow the distribution of natural images.
Therefore, the normal light pretrained model can also accomplish pretext tasks on enhanced low-light images.
Using this property, we fix both the backbone and the head, and train our deep concave curve by the self-supervised loss on enhanced low-light images.
Since semantics are taken care of by the MLP head, our deep concave curve will not be disturbed by \wj{it}.
Through training to improve pretext task performance, our model learns how to make the feature extractor better perceive the enhanced low-light image, therefore implicitly benefiting downstream tasks in low-light conditions.

\vspace{-1.5mm}
\subsection{Pretext Task Design}
\label{sec:pretext_task}

Now we need to find a pretext task that \wj{is able to} provide sufficient supervision for illumination adaptation.
Contrastive learning~\cite{MoCo} is one of the most popular self-supervised learning methods nowadays. However, \wj{it} adopts signal-related augmentations such as color jittering, which may destroy the model's perception of illumination.
Moreover, contrastive learning maximizes the feature distance between different samples, which may mislead the enhancement model \wj{into} \wj{generating} diverse but abnormal results.
Rotation prediction~\cite{Rotation} and jigsaw puzzling~\cite{Jigsaw} do not \wj{interfere} with the enhancement process, but they are not powerful enough, \wj{providing} limited supervision for illumination adaptation.

To effectively guide the enhancement model, we propose rotated jigsaw puzzles.
We first rotate the input image by random angles, then apply 3$\times$3 jigsaw shuffling and ask the network to recognize the permutation.
Compared with traditional jigsaw puzzling, rotation augmentation can increase the difficulty of the pretext task, which forces the MLP heads to better understand semantics, thus \wj{introduces} more supervision for adaptation.

The learning objective is as follows:
\begin{equation}
\label{eq:loss}
    \begin{aligned}
    \mathcal{L}_L & = \mathcal{L}_c(\hat{p}^{r-jig}_L, p^{r-jig}_L), \\
    \mathcal{L}_N & = \mathcal{L}_c(\hat{p}^{r-jig}_N, p^{r-jig}_N),
    \end{aligned}
\end{equation}
where $\mathcal{L}_c$ is the cross-entropy loss, $\hat{p}^{r-jig}$ is the predicted jigsaw permutation of the randomly rotated and jigsaw shuffled image, $p^{r-jig}$ is the ground truth permutation, $L$ stands for low light and $N$ stands for normal light.

\section{Framework Design and Analysis}
\vspace{-0.5mm}
\subsection{Unsupervised Low-light Adaptation}

With the above designs, we propose a new unsupervised low-light adaptation framework: self-aligned concave curve (\textbf{SACC}).
Given a pretrained normal-light downstream model, we use its feature backbone to train our deep concave curve with Eq.~(\ref{eq:loss}). While testing, we first enhance the input image and then apply the downstream model.
Different from existing low-light adaptation methods, SACC is blind to both normal and low-light annotations and \wj{does not even need to} adjust the downstream model.
\wj{Although only performing illumination enhancement, SACC achieves superior performance by taking full advantage of image data and machine features.}

Besides insufficient light, there are other low-light factors, such as color bias and noise.
For the former, SACC can handle color bias.
\todo{First, SACC has independent curves for each color channel.
Second, downstream feature extractors are usually trained on high-quality normal-light images without color bias, thus encouraging our model to restore balanced color.}
For the latter, real camera noise is quite hard to remove, but improving the model's robustness to noise is much easier.
On this basis, we adopt pseudo labeling to adapt the downstream model.
We predict labels of the enhanced low-light data, remove predictions of low confidence, and then fine-tune the downstream model on normal light data with labels and low-light data with pseudo labels.
Our strategy can not only implicitly handle noises but also narrow other kinds of domain gaps, such as scene differences.
This advanced version is called \textbf{SACC+}.

\vspace{-1.5mm}
\subsection{Analysis: Why Self-Supervised Alignment?}
\label{sec:analysis_alignment}

\begin{table}[t]
   \centering
   \small
   \caption{Effects of different learning strategies for guiding our illumination enhancement model.}
   \vspace{-1mm}
   \label{table:pretext}
    \begin{tabular}{l|l|c}
    \toprule
    Category                            & Strategy              & mAP (\%) \\ %
    \midrule
    Baseline                            & -                     & 16.09 \\ 
    \midrule
    \multirow{2}{*}{Discrepancy Metrics}   & CMD~\cite{CMD}        & 28.90 \\ 
                                           & MMD~\cite{MMD}        & 33.76 \\ 
    \midrule
    Adversarial Learning                & LSGAN~\cite{LSGAN}    & 38.05 \\ 
    \midrule
                        & MoCo~\cite{MoCo} & 38.46 \\ 
    Self-Supervised     & Rotation~\cite{Rotation} & 39.59\\ 
    Learning            & Jigsaw~\cite{Jigsaw} & 41.01 \\ 
                        & Rotated Jigsaw (proposed) & \textbf{41.31} \\ 
	\bottomrule
    \end{tabular}
    \vspace{-2.5mm}
\end{table}

In the following, we verify our self-supervised alignment strategy through experiments.
The analysis is based on DARK FACE~\cite{DARKFACE} and WIDER FACE~\cite{DARKFACE}.
\todo{The goal is to improve the performance of WIDER-FACE-pretrained DSFD~\cite{DSFD} on DARK FACE by enhancing input images.}
We do not use multiscale testing for simplicity.

Adapting from DARK FACE to WIDER FACE is especially hard \wj{since} WIDER FACE is collected from the Internet while DARK FACE is captured at nighttime streets.
Due to this huge scene gap, discrepancy metrics do not perform well as shown in Table~\ref{table:pretext}.
Adversarial learning leads to a better result, but the training is very unstable. In experiments, the balance between the low-light enhancement model and the discriminator often crashes.
In comparison, self-supervised learning is more effective and easier to train.
The contrastive learning method MoCo~\cite{MoCo} has been proved powerful for training networks from scratch.
But for adapting illumination, MoCo is less effective than rotation~\cite{Rotation} and jigsaw~\cite{Jigsaw}.
Finally, our rotated jigsaw strategy achieves the best result, which is in line with our motivation to increase the difficulty of the pretext task.




\vspace{-1.5mm}
\subsection{Analysis: Why Deep Concave Curve?}

\begin{table}[t]
  \centering
  \small
  \caption{Effects of different low-light enhancement backbones under our asymmetric self-supervised strategy.}
  \vspace{-1mm}
  \label{table:backbone}
    \begin{tabular}{l|l|c}
    \toprule
    Category    & Method   & mAP (\%) \\ %
    \midrule
    Baseline    & -         & 16.09 \\ 
    \midrule
    \multirow{3}{*}{Other}
    & EnlightenGAN~\cite{Enhance_EnlightenGAN} & 18.42 \\ 
    & Zero-DCE~\cite{Enhance_ZeroDCE}          & 25.36 \\ 
    & Gamma Correction $x^\gamma$ & 38.25 \\ 
    \midrule
    \multirow{4}{*}{Ours}
    & No Constraint         & 12.44 \\ 
    & $\nabla g \geq$ 0        & 34.40 \\ 
    & $\nabla g \geq$ 0 and $\nabla^2 g \leq$ 0 (proposed)      & \textbf{41.31} \\ 
    & $\nabla g \geq$ 0, $\nabla^2 g \leq$ 0, and $\nabla^3 g \geq$ 0      & 40.74 \\ 
    \bottomrule
    \end{tabular}
\end{table}

\begin{table}[t]
  \centering
  \small
  \caption{Comparison with other low-light enhancement networks under different training strategies.}
  \vspace{-1mm}
  \label{table:sa_enhancement}
    \begin{tabular}{l|l|c} 
    \toprule
    Method                      & Version               & mAP (\%) \\ %
    \midrule
    Baseline                    & -                     & 16.09 \\ 
    \midrule
                                & Original              & 28.85 \\ 
    EnlightenGAN~\cite{Enhance_EnlightenGAN} \ \ \ 
                                & Retrained             & 20.08 \\ 
                                & Original + Eq.~(\ref{eq:loss}) & 32.88 \\ 
    \midrule
                                & Original              & 38.32 \\ 
    Zero-DCE~\cite{Enhance_ZeroDCE}
                                & Retrained             & 32.53 \\ 
                                & Original + Eq.~(\ref{eq:loss})   & 40.54 \\ 
    \midrule
    SACC (Ours) & - & \textbf{41.31} \\ 
    \bottomrule
    \end{tabular}
    \vspace{-2mm}
\end{table}

Next, we raise another question: under the proposed training strategy, is deep concave curve superior to other network architectures?

\begin{figure}[t]
    \centering
  \includegraphics[width=0.99\linewidth]{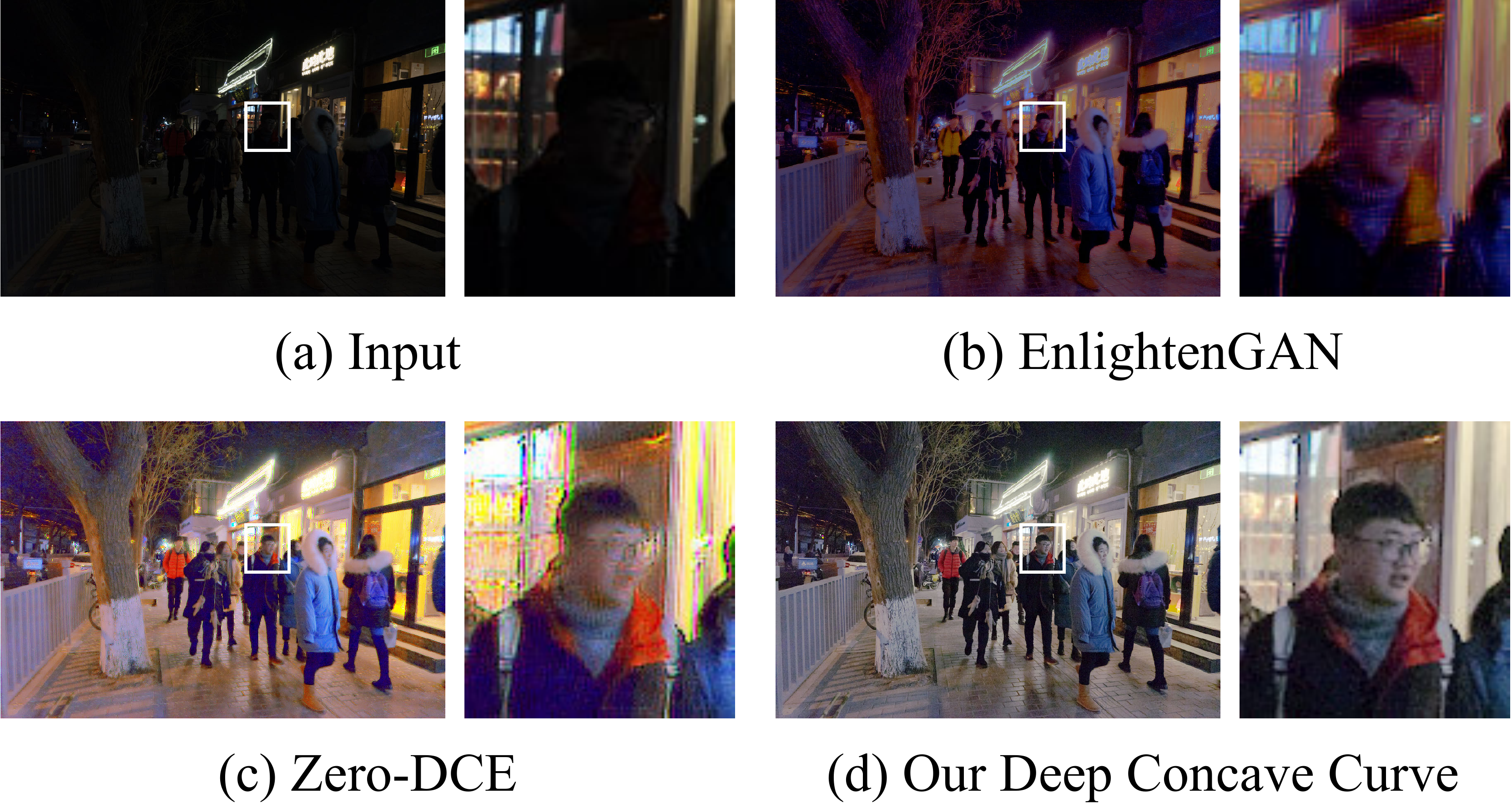}
  \vspace{-1mm}
  \caption{Effects of asymmetric self-supervised learning with different low-light enhancement model backbones.}
  \label{fig:backbone}
\end{figure}

We first explore two representative models: EnlightenGAN~\cite{Enhance_EnlightenGAN} and Zero-DCE~\cite{Enhance_ZeroDCE}.
EnlightenGAN adopts a widely-used U-net~\cite{UNet} like image-to-image architecture.
Zero-DCE adjusts images by element-wise second-order curves.
We retrain these two networks with Eq.~(\ref{eq:loss}). Other experimental settings are the same as Sec.~\ref{sec:analysis_alignment}.
As shown in Table~\ref{table:backbone}, their mAP scores are less than 30.
In Fig.~\ref{fig:backbone}(b) and (c), we can see that there are weird line artifacts.
As far as we are concerned, these lines are trivial solutions for the rotated jigsaw permutation problem.
In other words, EnlightenGAN and Zero-DCE get overfitted to Eq.~(\ref{eq:loss}), therefore cannot generalize to the downstream high-level vision task.
In comparison, our deep concave curve \wj{obtains} better visual results and mAP scores. \wj{This} is because our spatially sharing, monotonically increasing and concavity constraints act as regularization and prevent the model from carving cheating symbols or hints on images.


\wj{Na\"ive gamma correction $x^\gamma$ ($0$$<$$\gamma$$<$$1$) has been commonly used for low-light enhancement, but the adaptation performance is not as good as ours as shown in Table~\ref{table:backbone}.
This is because, for each color channel, $x^\gamma$ only has one learnable parameter $\gamma$. In contrast, our curve has 255 parameters.
Compared with $x^\gamma$, our curve has a higher degree of freedom and thus is more powerful.}

\begin{figure}[t]
    \centering
  \includegraphics[width=0.99\linewidth]{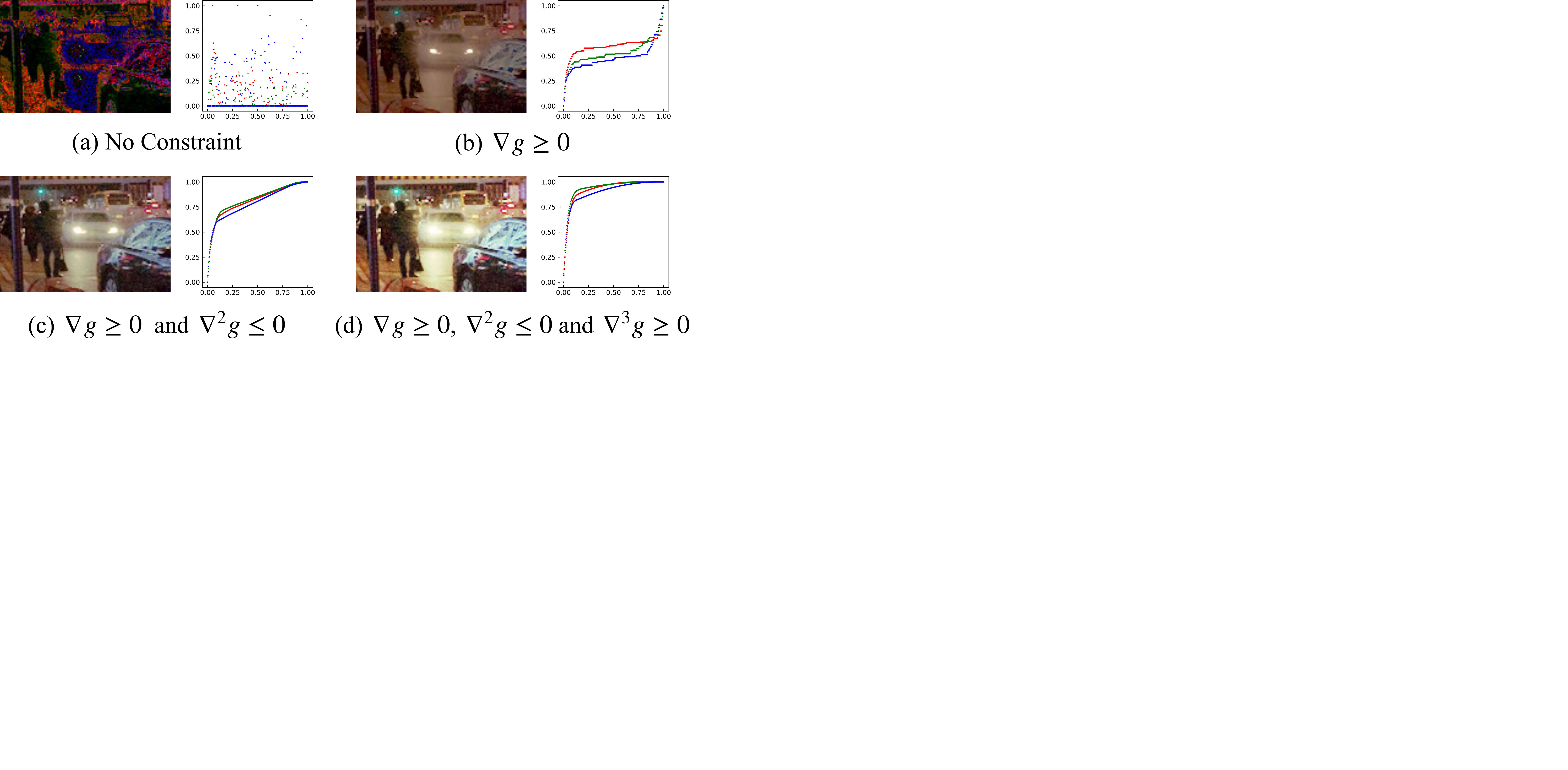}
  \vspace{-1mm}
  \caption{Low-light enhancement results (left) and curve shapes (right) under different curve form constraints.}
  \label{fig:diff_derivative}
  \vspace{-3mm}
\end{figure}

\wj{We further explore the necessity of each constraint in Table~\ref{table:backbone} and Fig.~\ref{fig:diff_derivative}.
Without any constraint, the enhanced results are far from natural images.
With only $\nabla g \geq$ 0, the curve does not look continuous and leads to low contrast and dull color as shown in Fig.~\ref{fig:diff_derivative}(b).
With $\nabla g \geq$ 0 and $\nabla^2 g \leq$ 0, the model achieves the best result.
We also try $\nabla^3 g \geq$ 0, but the mAP score gets slightly worse and the image gets partially over-exposed.
The reason might be that too strong constraints limit the expression ability of the model.
Also, integrating too many times makes the weights of the integral matrix extremely large, resulting in small gradients and increasing the difficulty of neural network training.}

Finally, we discuss: if the key lies in preventing overfitting to Eq.~(\ref{eq:loss}), will other `regularizations' also work?
We explore the constraints of adversarial learning in EnlightenGAN and a group of spatial, exposure, color, and smoothness non-reference loss functions in Zero-DCE.
As shown in Table~\ref{table:sa_enhancement}, when we retrain EnlightenGAN and Zero-DCE on DARK FACE and WIDER FACE, their performances get degraded, because their learning strategies are specially designed for their original datasets.
Then, we try to keep their loss functions on their original datasets, but add Eq.~(\ref{eq:loss}) on DARK FACE and WIDER FACE.
\wj{In Table~\ref{table:sa_enhancement}, we can see that the mAP scores are higher, demonstrating the effectiveness of Eq.~(\ref{eq:loss}) on other models.
This also indicates that adversarial learning and non-reference loss functions are effective `regularizations' as well.}
\wj{But even when using} Eq.~(\ref{eq:loss}), EnlightenGAN and Zero-DCE still perform worse than our deep concave curve.
\wj{This} is because adversarial and non-reference learning introduces extra conflicts against high-level vision, while our deep concavity is a purer regularization, \wj{and thus} can maximize the potential of Eq.~(\ref{eq:loss}).

\vspace{-2mm}
\subsection{Analysis: Denoising or SACC+?}

\begin{table}[t]
  \centering
  \small
  \caption{Effects of denoising enhanced low-light images and the proposed advanced version SACC+.}
  \vspace{-1mm}
  \label{table:denoise}
    \begin{tabular}{l|l|c} 
    \toprule
    Method                      & Version               & mAP (\%) \\ %
    \midrule
    Baseline                    & SACC                  & 41.31 \\ 
    \midrule
    \multirow{2}{*}{Denoising}
                                & SACC + BM3D~\cite{BM3D}             & 24.51 \\ 
                                & SACC + Neighbor2Neighbor~\cite{Neighbor2Neighbor}             & 40.54 \\ 
    \midrule
    Proposed                    & SACC+   & \textbf{45.51} \\ 
    \bottomrule
    \end{tabular}
    \vspace{-1mm}
\end{table}

We further discuss the necessity of denoising.
We evaluate two representative methods: the widely-used BM3D~\cite{BM3D} and the learning-based Neighbor2Neighbor~\cite{Neighbor2Neighbor}.
These two methods can remove noise from the perspective of human vision.
However, denoising also blurs details, which \wj{impedes} machine vision.
As a result, the accuracy of face detection degrades as shown in Table~\ref{table:denoise}.
In comparison, fine-tuning with low-light pseudo labels greatly improves the performance, supporting our design of SACC+.

\begin{figure}[t]
  \centering
  \includegraphics[width=0.99\linewidth]{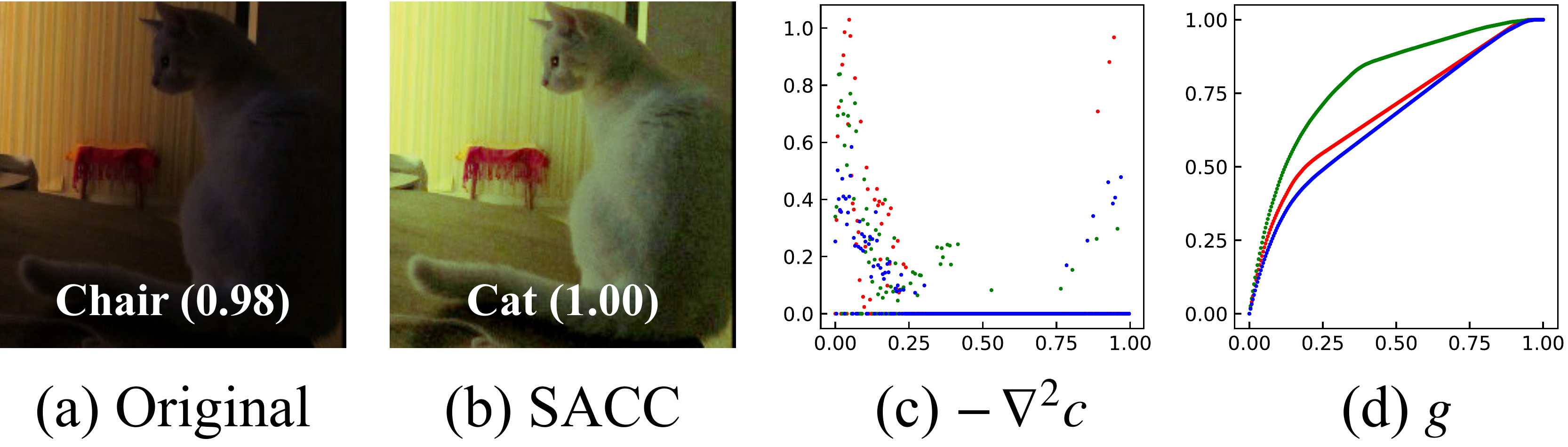}
  \vspace{-1mm}
  \caption{Example classification result. The original image is wrongly categorized as ``Chair'' with a confidence of 0.98, after enhancement, it can be correctly recognized as ``Cat''.}
  \label{fig:enh_example}
  \vspace{-3mm}
\end{figure}

\vspace{-1mm}
\section{Experiments}

In this section, we provide implementation details, benchmarking results, and more experimental analysis and applications.

\vspace{-2mm}
\subsection{Implementation Details}
\label{sec:exp_setting}

\wj{Our deep concave curve can be applied to various models and vision tasks.
In this paper, we show the results of ResNet~\cite{ResNet}, DSFD~\cite{DSFD} (VGG~\cite{VGG} backbone), I3D~\cite{I3D} (3D-CNN), and PWC-Net~\cite{PWCNet} (feature pyramid backbone), involving four different visual tasks.}



For SACC, we first \new{pretrain} the pretext task head on normal light images with the feature extractor fixed.
Then, we fix the pretext task and train our deep concave curve on low light images.
For SACC+, we first fix the \new{pretrained} SACC and then fine-tune the downstream model on normal-light and SACC-enhanced low-light data.
Due to page limit, detailed experimental settings for each task 

\noindent are provided in the supplementary material.


\begin{table}[t]
  \centering
  \small
  \caption{Adaptive low-light classification comparison results. $\dagger$ denotes that the low-light enhancement model is retrained.}
  \vspace{-1mm}
  \label{table:codan}
    \begin{tabular}{l|l|c}
    \toprule
    Category    & Method                    & Top-1 (\%) \\
    \midrule
    Baseline    & ResNet-18~\cite{ResNet}   & 55.04 \\
    \midrule
    Fully Supervised	& Finetuned ResNet-18~\cite{ResNet} & 71.52 \\
    \midrule
                & RetinexNet~\cite{Enhance_RetinexNet} 		& 44.72 \\
                & Zero-DCE$\dagger$~\cite{Enhance_ZeroDCE}  & 50.80\\ 
                & EnlightenGAN~\cite{Enhance_EnlightenGAN} 	& 57.76 \\
                & Zero-DCE++~\cite{Enhance_ZeroDCE++}	 	& 58.56 \\
                & MF~\cite{Enhance_MF}                      & 59.20 \\   
    Low-Light   & KinD~\cite{Enhance_KinD} 					& 59.28 \\
    Enhancement & LIME~\cite{Enhance_LIME}                  & 59.44 \\
                & Zero-DCE~\cite{Enhance_ZeroDCE}           & 59.44 \\
                & RUAS$\dagger$~\cite{RUAS}                 & 59.84 \\
                & EnlightenGAN$\dagger$~\cite{Enhance_EnlightenGAN}	& 60.24 \\
                & LLFlow~\cite{LLFlow} & 60.72 \\
                & \textbf{SACC} (Ours)                      & \textbf{61.44} \\
    \midrule
                    & CMD~\cite{CMD}                & 55.92 \\
    Unsupervised    & AdaBN~\cite{AdaBN}            & 59.68 \\
    Domain          & DANN~\cite{DANN}              & 59.76 \\
    Adaptation      & CIConv~\cite{CIConv}          & 60.96 \\
                    & \textbf{SACC+} (Ours)         & \textbf{63.92} \\
	\bottomrule
    \end{tabular}
    \vspace{-8mm}
\end{table}

\begin{figure*}[t]
  \centering
  \includegraphics[width=0.99\linewidth]{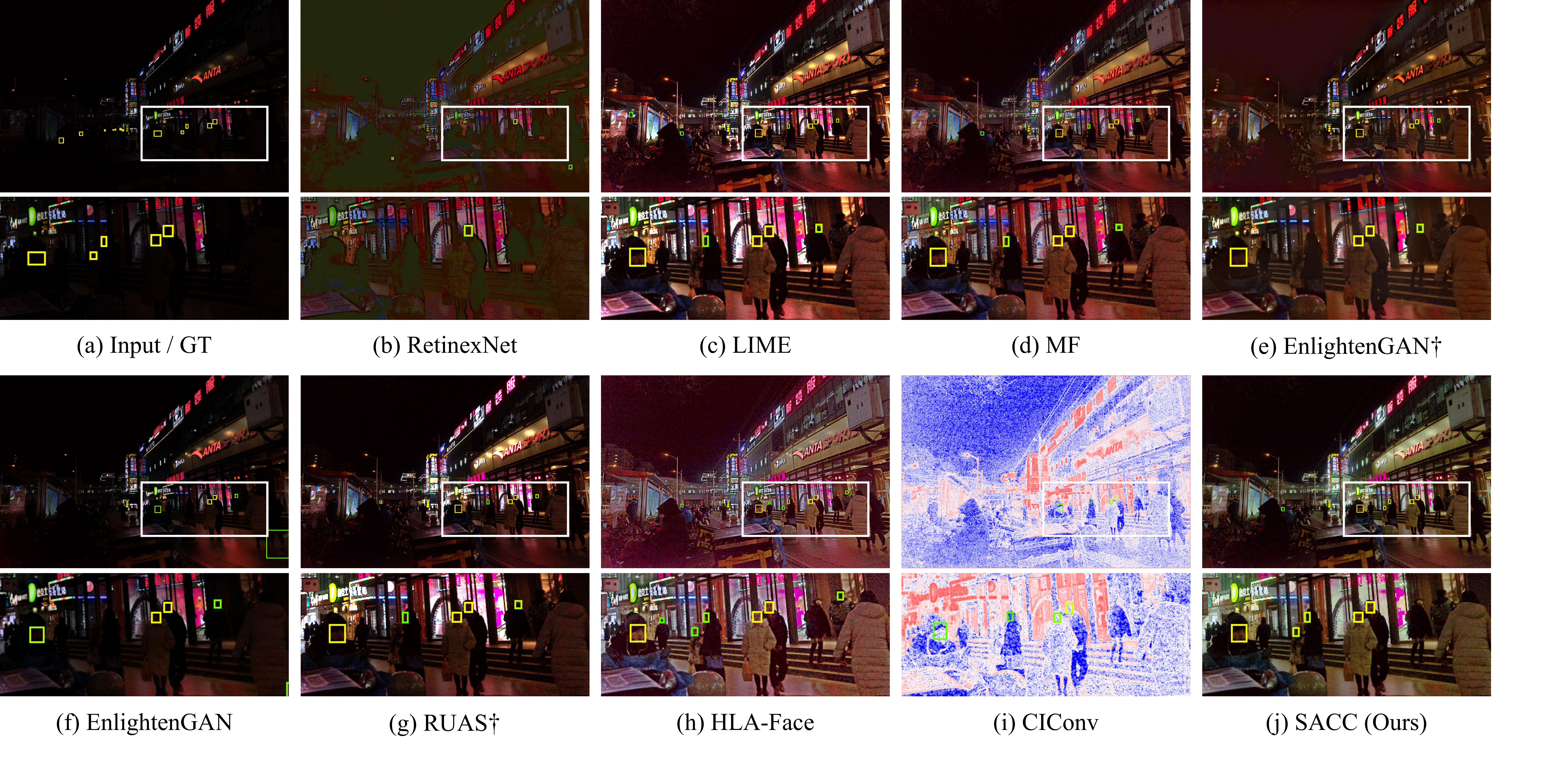}
  \vspace{-1mm}
  \caption{Subjective comparison results for dark face detection.
  $\dagger$ denotes that the low-light enhancement model is retrained. The color of the bounding boxes represents the confidence of recognition, with yellow indicating higher confidence.}
  \label{fig:comp_enh}
  \vspace{-1mm}
\end{figure*}

\vspace{-2mm}      
\subsection{Low-light Object Classification}   

\wj{We first evaluate our approach by one of the most fundamental vision tasks: image classification.
CODaN~\cite{CIConv} is a 10-class dataset for low-light adaptation. It contains 13k daytime and 2.5k nighttime images.} We use its official daytime setting and split 1.25k nighttime images for training/validation and the others for testing. The task is to adapt ResNet-18~\cite{ResNet} from daytime to nighttime.
Our method is compared with nine low-light enhancement and four unsupervised domain adaptation methods.
\todo{Enhancement methods only adjust input images, while adaptation methods also adjust the classifier.
They all cannot obtain low-light labels.
We additionally show the result of fully supervised learning as a reference of performance upper bound, where the baseline is trained with low-light labels.}

The results are shown in Table~\ref{table:codan}.
Low-light enhancement methods restore input low-light images from the perspective of human visual quality and \wj{neglect} machine vision, \wj{thus resulting} in limited performance on CODaN.
Unsupervised domain adaptation methods adapt the classifier to nighttime.
CMD~\cite{CMD}, AdaBN~\cite{AdaBN}, and DANN~\cite{DANN} are designed for normal light, but the day-night domain gap is more complex.
Accordingly, their accuracies are all below 60\%.
CIConv~\cite{CIConv} uses a color invariant layer.
Benefitting from its robust feature, CIConv is superior to other low-light enhancement and unsupervised domain adaptation methods.
But compared with our methods, CIConv is less effective.
Through adjusting illumination from the perspective of machine vision, our SACC best adapts the daytime classifier to nighttime.
With the help of pseudo label fine-tuning to adjust the downstream classifier, the accuracy of our SACC+ is even higher.
\new{Fig.~\ref{fig:enh_example} shows an example result. $-\nabla^2 c$ is very discrete, but after integral $g$ appears like a continuous curve.
As the input image suffers from red color bias, our model brightens more on the green channel, which adjusts the color balance.}

\begin{table}[t]
   \centering
   \small
   \caption{Adaptive dark face detection comparison results. $\dagger$ denotes that the low-light enhancement model is retrained.}
   \vspace{-1mm}
   \label{table:darkface}
    \begin{tabular}{l|l|c}
    \toprule
    Category            & Method & mAP (\%) \\ 
    \midrule
	Baseline            & DSFD~\cite{DSFD}						    & 16.09 \\ 
    \midrule
	Fully Supervised	& Finetuned DSFD~\cite{DSFD}			    & 45.99 \\ 
	\midrule
    & RetinexNet~\cite{Enhance_RetinexNet} 		& 12.04 \\ 
    & KinD~\cite{Enhance_KinD} 					& 15.84 \\ 
    & EnlightenGAN$\dagger$~\cite{Enhance_EnlightenGAN}	& 20.77 \\ 
    & EnlightenGAN~\cite{Enhance_EnlightenGAN} 	& 31.31 \\ 
    & Zero-DCE $\dagger$~\cite{Enhance_ZeroDCE}	& 37.30 \\ 
    Low-Light
    & LLFlow~\cite{LLFlow} & 37.41 \\ 
    Enhancement
    & RUAS$\dagger$~\cite{RUAS}                          & 38.36 \\ 
    & LIME~\cite{Enhance_LIME} 					& 40.71 \\ 
    & Zero-DCE++~\cite{Enhance_ZeroDCE++}	 	& 40.90 \\ 
    & Zero-DCE~\cite{Enhance_ZeroDCE}			& 41.27 \\ 
    & MF~\cite{Enhance_MF}						& 41.43 \\ 
    & \textbf{SACC} (Ours)                      & \textbf{44.57} \\ 
	\midrule
		                & CIConv~\cite{CIConv}                      & \textcolor{white}{0}4.40 \\ 
	\multirow{2}{*}{Unsupervised}        & OSHOT~\cite{DA_DInnocenteBBCT_ECCV20} 	& 25.38 \\ 
	\multirow{2}{*}{Domain}              & Progressive DA~\cite{DA_WACV20}		    & 28.47 \\ 
	\multirow{2}{*}{Adaptation}          & Pseudo Labeling~\cite{DA_Cross_CVPR18}    & 35.07 \\ 
	                    & HLA-Face~\cite{HLAFace}                   & 44.44 \\
	                    & \textbf{SACC+} (Ours)        & \textbf{46.91} \\ 
	\bottomrule
    \end{tabular}
    \vspace{-3mm}
\end{table}

\vspace{-1.5mm}
\subsection{Dark Face Detection}

We further conduct a benchmark on unsupervised dark face detection.
WIDER FACE~\cite{WIDERFACE} contains 32k images of various events and scenes.
DARK FACE~\cite{DARKFACE} contains 10k nighttime street scene images.
We use their official splittings.
The baseline is DSFD~\cite{DSFD}.

Results are shown in Table~\ref{table:darkface}.
\wj{Although RetinexNet~\cite{Enhance_RetinexNet} and KinD~\cite{Enhance_KinD} consider detail reconstruction and noise reduction, they introduce extra artifacts for machine vision} and thus degrade the performance of face detection.
Other low-light enhancement methods can \wj{benefit DSFD}, but their mAP scores are lower than ours.
\wj{These methods may make enhanced images look unnatural in machine vision, which can mislead the face detector as shown in Fig.~\ref{fig:comp_enh}.}

\begin{figure*}[t]
    \centering
  \includegraphics[width=0.99\linewidth]{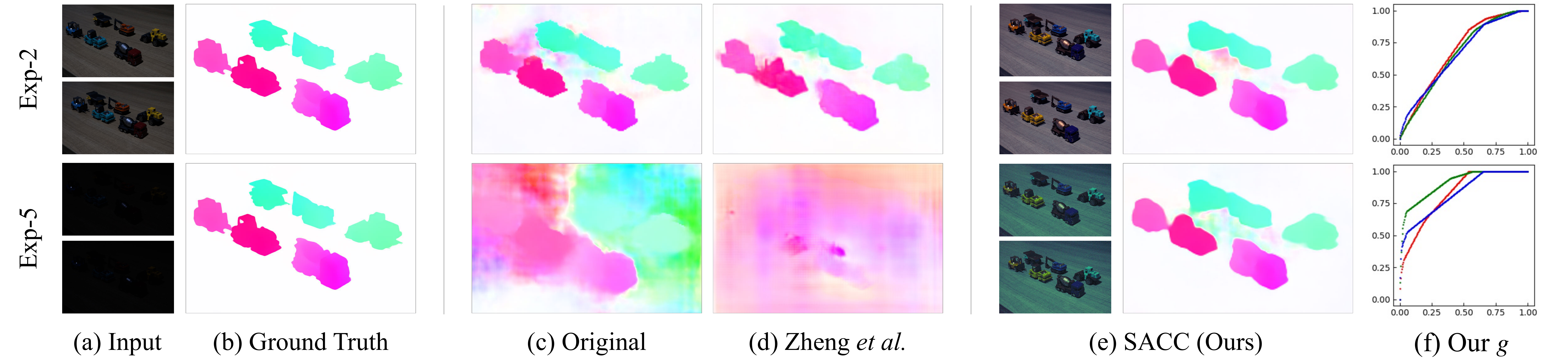}
  \vspace{-1mm}
  \caption{Optical flow estimation results of the same scene under different illumination levels.}
  \label{fig:comp_flow}
\end{figure*}

Unsupervised domain adaptation methods OSHOT~\cite{DA_DInnocenteBBCT_ECCV20}, Progressive DA~\cite{DA_WACV20}, and Pseudo Labeling~\cite{DA_Cross_CVPR18} cannot effectively handle the normal-low light gap.
Although CIConv~\cite{CIConv} is effective on CODaN, due to the much lower illumination of nighttime street scenes, the color invariant layer of CIConv fails on DARK FACE.
As visualized in Fig.~\ref{fig:comp_enh}(i), the representation of CIConv suffers from severe noise.
HLA-Face~\cite{HLAFace} is the state-of-the-art for WIDER-DARK unsupervised adaptation. 
It adopts a complex low-level and high-level joint adaptation strategy.
In comparison, our SACC only adjusts illumination but still outperforms HLA-Face.
Our advanced version SACC+ even outperforms the fully supervised upper bound, demonstrating \wj{the great} potential of our method.


\wj{Now we analyze the generalization of DSFD-trained SACC to other face detectors: PyramidBox~\cite{PyramidBox} and MogFace~\cite{MogFace}.
As shown in Table~\ref{table:more_results}, our performance is consistently good even for unseen downstream models, indicating that high-level vision knowledge is universal among different machine models.}

\begin{table}[t]
  \centering
  \small
  \caption{More results on DARK FACE~\cite{DARKFACE}. \best{Red} denotes the best and \second{blue} denotes the second best performance.}
  \vspace{-1mm}
    \label{table:more_results}
    \begin{tabular}{l|ccc|c|cc}
    \toprule
    & \multicolumn{4}{c|}{Detection mAP (\%)} & \multicolumn{2}{c}{NR-IQA} \\
    \cmidrule(lr){2-5} \cmidrule(lr){6-7} 
    & \cite{PyramidBox}
    & \cite{DSFD}
    & \cite{MogFace}
    & AVG
    & \cite{NIQE}
    & \cite{SSEQ} \\
    \midrule
    Original     		                                & 13.99 & 16.09 & 16.36 & 15.48 & \ 8.5 & 24.1 \\
    \midrule
    RetinexNet~\cite{Enhance_RetinexNet} 		        & 11.42 & 12.04 & 14.56 & 12.67 & \ 7.7 & 21.9 \\
    KinD~\cite{Enhance_KinD} 			                & 15.61 & 15.84 & 21.27 & 17.57 & \ 9.8 & 42.2 \\
    EnlightenGAN$\dagger$~\cite{Enhance_EnlightenGAN}	& 19.54 & 20.77 & 24.02 & 21.44 & \ 9.0 & 22.6 \\
    EnlightenGAN~\cite{Enhance_EnlightenGAN}            & 28.45 & 31.31 & 35.79 & 31.85 & \ 9.7 & 14.1 \\
    Zero-DCE$\dagger$~\cite{Enhance_ZeroDCE}	        & 33.41 & 37.30 & 37.75 & 36.15 & \ 6.7 & 18.4 \\
    LLFlow~\cite{LLFlow}                                & 32.84 & 37.41 & 41.08 & 37.11 & \ 8.4 & 10.3 \\ 
    RUAS$\dagger$~\cite{RUAS}                           & 32.77 & 38.36 & 40.71 & 37.28 & \ \best{6.2} & \ \ \second{4.4} \\
    LIME~\cite{Enhance_LIME} 			                & 35.69 & 40.71 & 42.82 & 39.74 & \ 6.5 & \ \ 5.7 \\
    Zero-DCE++~\cite{Enhance_ZeroDCE++}  	 	        & 35.56 & 40.90 & 43.45 & 39.97 & \ 6.4 & \ \ 6.7 \\
    Zero-DCE~\cite{Enhance_ZeroDCE}		                & 35.95 & 41.27 & 43.62 & 40.28 & \ 6.4 & \ \ 7.9 \\
    MF~\cite{Enhance_MF}				                & \second{37.49} & \second{41.43} & \second{43.87} & \second{40.93} & \ 6.5 & \ \ 8.2 \\
    \textbf{SACC} (Ours)           & \best{39.20} & \best{44.57} & \best{46.45} & \best{43.41} & \ \second{6.3} & \ \ \best{3.2} \\ 
    \bottomrule
    \end{tabular}
    \vspace{-3mm}
\end{table}

\subsection{Low-light Action Recognition}

\begin{table}[t]
  \centering
  \small
  \caption{Video action recognition comparison results.}
  \vspace{-1mm}
  \label{table:ARID}
    \begin{tabular}{c|ccc}
    \toprule
    \ Original\ \   & \ \  StableLLVE~\cite{StableLLVE} & SMOID~\cite{SMOID} & \textbf{SACC} (Ours) \\ 
    \midrule
    47.14\%  & 49.70\% & 50.18\% & \textbf{52.13\%} \\
	\bottomrule
    \end{tabular}
\end{table}

\wj{Although initially designed for images, our method can also generalize to video tasks.
Here we evaluate our SACC with 11-class low-light video action recognition.
We gather about 800 low-light video clips from ARID~\cite{ARID}.
Normal light data consists of 2.6k normal light video clips from HMDB51~\cite{HMDB51}, UCF101~\cite{UCF101}, Kinetics-600~\cite{Kinetics}, and Moments in Time~\cite{MiT}.
The action recognizer is I3D~\cite{I3D} based on 3D-ResNet~\cite{3D_ResNet}.
We report the top-1 accuracy.}


\todo{We extend SACC to videos by merging all frames into a big image, predicting a unified $g$ for the whole video clip, then applying $g$ to each frame. Since $g$ is temporally shared, temporal consistency can be naturally maintained.}

\wj{In Table~\ref{table:ARID}, video enhancement methods StableLLVE~\cite{StableLLVE} and SMOID~\cite{SMOID} can benefit action recognition but attain limited performance gain.
Our SACC instead increases the accuracy by about 5\%, showing the superiority of our method for videos.}


\subsection{Optical Flow Estimation in the Dark}

\begin{table}[t]
  \centering
  \small
  \caption{Adaptive low-light optical flow estimation results.}
  \vspace{-1mm}
  \label{table:VBOF}
    \begin{tabular}{l|cccc|c}
    \toprule
                                & Exp-2 & Exp-3 & Exp-4 & Exp-5 & AVG \\
    \midrule
    Original                    & 10.21	& 11.90	& 14.36	& 17.82 & 13.57 \\
    \midrule
    Zheng \etal~\cite{VBOF}     & 9.04 & 8.81 & 8.78 & 9.31 & 8.99 \\
    \textbf{SACC} (Ours)        & \textbf{6.70}	& \textbf{7.03}	& \textbf{7.57}	& \textbf{8.47} & \textbf{7.44} \\
	\bottomrule
    \end{tabular}
\end{table}

Last but not least, we explore a task that differs a lot from classification and detection: optical flow estimation.
The VBOF~\cite{VBOF} dataset contains 10k image pairs under various brightness levels.
We select Exp-2 to Exp-5 subsets as the adaptation target domain, where Exp-5 is the darkest.
The ground truth flow fields are originally estimated by FlowNet 2.0~\cite{FlowNet2} on the sharp and normal light Exp-1 subset, we re-calculate them with a more state-of-the-art approach GMA~\cite{GMA}.
The baseline to adapt is PWC-Net~\cite{PWCNet}.
Performance is evaluated by end-point error (EPE).

Zheng~\etal~\cite{VBOF} proposed to adapt optical flow models by simulating the noise of dark images and synthesizing a low-light training dataset.
However, the synthesis process only considers signal distribution and neglects machine vision, \wj{and thus} the performance gain is worse than ours as shown in Table~\ref{table:VBOF}.
Furthermore, as shown in Fig.~\ref{fig:comp_flow}, our SACC is robust to \wj{different} input illumination levels.

\subsection{Subjective Human Vision}
\label{sec:subjective}




We further analyze the relationship between human and machine vision.
We benchmark two non-reference image quality assessment (NR-IQA) scores: NIQE~\cite{NIQE} and SSEQ~\cite{SSEQ} in Table~\ref{table:more_results}.
Although our SACC only considers high-level tasks, our NR-IQA scores still rank high among all methods. As shown in Fig.~\ref{fig:comp_enh}, our enhanced result \wj{has} less artifacts and better color balance.

The correlation coefficient of NIQE/SSEQ and the average mAP of all detectors is -0.69/-0.84.
This experiment indicates that there is indeed a correlation between human and machine vision, but good visual quality does not always guarantee good high-level performance, which validates our motivation \wj{for} designing enhancement models tailored for machine vision.

\section{Conclusion}

Towards high-level vision, we propose deep concave curve, which migrates illumination from low to normal light through a self-supervised strategy.
Our model architecture and learning strategy cooperate with each other, which form a powerful normal-to-low light adaptation framework.
Experimental results demonstrate the superiority of our framework.

\bibliographystyle{ACM-Reference-Format}
\bibliography{final.bib}

\end{document}